\documentclass[10pt, a4paper, twocolumn]{article}
\usepackage[a4paper,margin=2cm]{geometry}
\usepackage[utf8]{inputenc}
\usepackage[polutonikogreek, english]{betababel} 
\usepackage{microtype} 
\usepackage{amsmath,amsfonts,amsthm} 
\usepackage{hyperref}

\begin{document}

\title{Is Epicurus the father of Reinforcement Learning?} % The article title
\author{Eleni Vasilaki \vspace{5mm}\\
Department of Computer Science, University of Sheffield, United Kingdom\\ e.vasilaki@sheffield.ac.uk}
\maketitle

\begin{abstract}
The Epicurean Philosophy is commonly thought as simplistic and hedonistic. Here I discuss how this is a misconception and explore its link to Reinforcement Learning. Based on the letters of Epicurus, I construct an objective function for hedonism which turns out to be equivalent of the Reinforcement Learning objective function when omitting the discount factor.  I then discuss how Plato and Aristotle 's views that can be also loosely linked to Reinforcement Learning, as well as their weaknesses in relationship to it.  Finally, I emphasise the close affinity of the Epicurean views and the Bellman equation. 
\end{abstract}

%----------------------------------------------------------------------------------------
%	ARTICLE CONTENTS
%----------------------------------------------------------------------------------------

%\section{Section}
\vspace{5mm}
\latintext  When we think of Greek Philosophy, for most this constitutes the names Plato and Aristotle. This is not a surprise given the depth  and the amount of their work, which, for the largest part, remains intact till today. In the famous painting of Raphael, Plato and Aristotle are portrayed right in the middle, dominating the scene. It is hard to pay attention to the young man at the bottom left-hand corner, who is said to be the philosopher Epicurus. 

Epicurus is perhaps the most misunderstood philosopher of the Ancient world. Those who have heard his name before associate it with pleasure. Epicurus is commonly perceived as a hedonist, as someone who only promoted the immediate pleasures of the flesh. This, however, is a superficial and largely misleading presentation of his philosophy. 

Born in the island of Samos in 341BC \cite{Britannica,SEP}, also birthplace of the famous mathematician Pythagoras, Epicurus consequently lived in Athens, Lesbos, and Asia Minor. He was aware of the Aristotelian philosophy, which was a mainstream philosophical movement at the time, however he was dissatisfied with this view of the world and adopted instead the earlier view of Democritus:\\

 \textgreek{((>arq`ac e\~>inai  t\~wn <'olwn >at'omouc ka`i ken'on, t`a d' >'alla p'anta nenom'isjai))}  \cite{Democritus}\\

{\bf  ``The beginning of everything are atoms and space, everything else is in your mind.''}\\

Much of his philosophy builds upon the philosophy of Democritus, complementing ideas that Aristotle had vigorously criticised. With his views not being very popular in Athens, where the Aristotelian philosophers dominated, Epicurus moved to Lesbos. However, his philosophy was again not accepted; even worse Epicurus was accused of serious crimes and had to escape leaving the island in stormy weather, to reach the Asia Minor coast.

At the time, there were several Greek cities-colonies in Asia Minor. These cities were considered more progressive than the mainland cities; a suitable environment for the Philosophy of Epicurus.  For instance it was not uncommon for the women of the colonies to get an education, in fact Asia Minor was the birthplace of many famous hetairas who are remembered by history. Hetairas are thought to be highly educated, high-end courtesans, companions of philosophers.  One such example is the eminent Aspasia of Miletus, companion of Pericles who was the most renowned Athenian leader in the 5th BC century, the golden age of the Ancient Greek world. Aspasia is said to have written Pericles' most inspiring speeches. 

If there was any place in the Ancient world that could have accepted the revolutionary Epicurean philosophy, this was Asia Minor.  Indeed Epicurus established a famous school in the city of Lampsacus \cite{Britannica} which, similar to the school of Pythagoras, accepted women and slaves among the students, something unacceptable at the elitist schools of Athens at the time. To put the matter in some perspective, Girton College in Cambridge was established as late as 1869AC to accept female students \cite{Girton}. Epicurus, similar to Pythagoras, accepted students based on merit, not gender or status. He was even accused of praising his female student Themista, wife of Leonteus more than worthy male philosophers. Themista, according to one source \cite{Britannica}, had previously been a hetaira. In 307BC, Epicurus, having gathered several followers returned once again in Athens, where he died in his early 70s \cite{Britannica,SEP}.

The world of Epicurus, similar to Democritus, consists of a world of interacting atoms, which account for the perceived properties of bodies. According to Epicurus, perception can be explained based on the ``interaction of atoms with the sense-organs'' \cite{IEP}. He added to the atomist theory the notion that ``all perception is true'' and that we can learn to pursue natural pleasures rather than misleading desires imposed by society \cite{SEP}. This view is much compatible to the general framework of Reinforcement Learning, where an agent interacts with its environment and receives feedback from it. It is reflected even closer by a most elaborated version of the agent-environment interaction proposed by Andy Barto and colleagues \cite{Barto}, according which the external world provides ``sensations'', that are internally interpreted by the agent (or organism) as reward signals, and that together with the perceived internal states will lead to decisions and actions. In the Epicurean philosophy, there is a clear mapping of good as pleasure, or in the Reinforcement Learning terminology positive reward, and evil as pain, or negative reward, punishment. There is also a clearly defined objective function:\\

 \textgreek{((...t`hn <hdon`hn >arq`hn ka`i t'elos l'egomen e\~>inai to\~u makar'iwc z\~hn))}  \cite{Epicurus} \\\

{\bf ``We say that pleasure is the beginning and the end of a happy life''},\\

\noindent Epicurus states in his letter to Menoeceus \cite{EpicurusEN}. 

However, pleasure in the Epicurean philosophy does not merely have the notion of an immediate reward, as perhaps misinterpreted by the critics (or the followers!) of Epicurus. Instead, it has the notion of a function to be maximised given future rewards or punishments, as the consequence of present actions. In the simplest form, we can write pleasure as a function R, parametrised by time t, which consist of a finite series of rewards of the future r at time t+1, t+2, t+3...T:

\begin{equation*}
R_t= r_{t+1}+r_{t+2}+r_{t+3}+\ldots+r_{T}
\end{equation*}

As in Reinforcement Learning \cite{SuttonBarto}, the reward r, in this context, can be positive, zero or negative, capturing also potential neutral or negative effects of an action. According to Epicurus, choices of actions depend on future long-lasting pleasures:\\ 

{\bf ``For we know pleasure as our first and inborn good, and we have it as the beginning of our every choice and avoidance, and we turn to it in using sensation as a standard for judging every good. And since pleasure is our first and natural good, for this reason we do not choose every pleasure, but there are times when we pass over many pleasures, when greater difficulty would result from them for us, and we consider many pains to be better than pleasures, whenever a greater and long-lasting pleasure will follow for us after we have suffered the pains. So every pleasure is a good through being naturally agreeable, nevertheless not every pleasure is to be chosen; just as it is also the case that every pain is bad, but not every pain is always by its nature to be avoided.''} (Translation taken from \cite{EpicurusEN}).\\

In this context, the word ``pleasure'' has two different meanings, one referring to immediate pleasures (or pains), and one referring to the general concept of pleasure, the objective function that is captured by $R_t$. In the same letter, the Philosopher offers advices regarding the future consequences of the present actions. This is equivalent of thinking that the Philosopher has obtained, by his experience and observation, the state-action values Q(s,a) defined in Temporal-Difference Reinforcement Learning, which expresses the total reward that can be collected by an agent in state s when taking action a. This prior knowledge of the Q-values might save others the time and effort that would be required to discover if an action is beneficial or harmful for their ``souls'', or which types of pleasure should be pursued:\\

{\bf ``For continual drinking and partying, or taking one's enjoyment of boys and women [...] do not produce a pleasant life, but sober reasoning which both examines the basis for every choice and avoidance and drives out the opinions which cause very great turmoil to take hold of our souls.''} (Translation taken from \cite{EpicurusEN}).\\

This is precisely the turning point for all the presumed ideas of a simplistic hedonism in the Epicurean philosophy. It is not just the pleasure of now that one has to be concerned with, but its future consequences. Drinking, for instance, would have an immediate positive reward at time t+1 but a negative reward at some other time in the future (i.e., hangover!). In general, the values of the actions in the Epicurean Philosophy seem independent of the state of the agent, as they capture rather general concepts. For instance, Epicurus valued friendship beyond other pleasures and advised against pursuing romantic love too much as this often brings jealousy and pain in the long run. Considering friendship investment as an action, its value is independent of the state of the agent (human) and higher than the value of the action of pursuing romantic love. In fact, to the best of my knowledge, Epicurus never got married. He lived a long life, and a happy one to the end, despite the difficulties that he faced. Epicurus, in his letters, advised that death is nothing to be afraid of as when we are here, death is not present, and when death is present, we do not exist  \cite{EpicurusEN}.

The evaluation of an action based on the addition of the consequent punishments and rewards is not unique in the Epicurean philosophy.  Plato had earlier suggested much the same. In Protagoras, Socrates argues:\\  

{\bf ``Then your idea of evil is pain, and of good is pleasure. Even enjoying yourself you call evil whenever it leads to the loss of a pleasure greater than its own, or lays up pains that outweigh its pleasures.''} (Translation taken from \cite{Plato})\\

Further in the text, however, Socrates claims that it is difficult to measure good and evil, it requires a special skill, an art or science, and Protagoras agrees with him. This claim goes against the idea of Reinforcement Learning, where a naive agent can learn a task with a trial and error approach (exploration/exploration). 

In Reinforcement Learning, the reward function is formulated identical to the pleasure function we defined above for finite time.  In the case of an infinite series of future rewards, it includes a discount factor as a mathematical necessity. The discount factor also captures the fact that an immediate reward is preferable to a future reward of the same size: we would rather have a million pounds now than in five years time. I have not found the notion of discounting future rewards in the Epicurean philosophy, and for this, the pleasure function is defined as a finite series. This simplification, however, does not affect the importance of the Epicurean philosophy, which can be viewed as a sophisticated reward maximisation practice.

Why is Epicurus, with his progressive thinking, only at the corner of Raphael's painting, and in the mind of most merely a hedonist? For one, most of his work is not available; what we know of him today is mainly due to the letters he wrote (Letter to Herodotus, Menoeceus, Pythocles) \cite{Britannica,SEP}. The prolific writing of Plato and Aristotle also mesmerised academics and didn't let much space for appreciation of the Epicurean philosophy available via limited sources. There may be one more reason though; Plato with his theory of Ideas (Forms) and Aristotle with his notion of the final cause \cite{SEP} offered a framework very much compatible with the existence of a God. Epicurus with his Democritean view of the world did not impose Gods as an implicit yet strong necessity of his philosophy. On the contrary, he believed that the soul died with the body and that he could disprove the notion of afterlife \cite{SEP}.

Is Epicurus the father of Reinforcement Learning? Perhaps most, including the author, would save the term for Rich Sutton and Andy Barto who wrote the bible of Reinforcement Learning \cite{SuttonBarto}. Indeed, in the Epicurean philosophy there is no algorithmic procedure for learning the action values similar to Temporal Difference Reinforcement Learning. Instead, the Philosopher suggests to use reason and experience \cite{12GreeksRomans} to choose the best action in terms of immediate and future pleasure, avoiding these that can cause great ``turmoil'' in the future. In this sense, the Epicurean model does not include a sufficiently detailed exploration/exploitation approach as in Reinforcement Learning, and yet it does state that sensation can be used to judge every good \cite{EpicurusEN}.  On the contrary, Plato considers the idea of knowing  good and evil a difficult matter that requires specialised  expertise \cite{Plato}, while Aristotle, similar to Epicurus \cite{12GreeksRomans}, appears to have embraced learning by trial and error \cite{OnAristotle} with his following statement, loosely linked to Reinforcement Learning: \\

 {\bf ``For the things we have to learn before we can do them, we learn by doing them''} (Translation taken from \cite{Aristotle}).\\

\noindent Aristotle, however, believed in an intrinsic teleology, explaining nature and objects via the purpose that they serve. In the context of his {\it Metaphysics} \cite{ArtistotleMataphysics}, this idea might suggest that brains have been designed for a specific purpose. To Aristotle, God is ``the ultimate cause and explanation of rational change and order in nature'' \cite{ArtistotleMataphysics}, while Epicurus explained physical phenomena in a proto-Darwinian way, as ``a process of natural selection'' \cite{IEP}, a truly advanced idea for his time. The Epicurean understanding of the world implies that brains have been evolved rather than designed to purpose. 

Epicurus' view of pleasure appears to be tightly related to the concept of Dynamic Programming, a key ingredient of Temporal Difference Reinforcement Learning. In particular, it reminds of the Bellman optimality equation, which determines that the unique, optimal policy of choosing actions is greedy (i.e., choose the action with the highest Q-value) yet assumes that Q-values are known or can be explicitly calculated \cite{SuttonBarto}. As such, the notion of the Reward function in Reinforcement Learning, as a finite sum of present and future rewards is perfectly and, perhaps, surprisingly captured by the philosophy of Epicurus.

\vspace{0.5cm}

\noindent {\bf \Large Acknowledgements}
\vspace{0.3cm}\\
The following text is largely based on my talk at the Sheffield Machine Learning Group Retreat, on the 2nd June 2017. Reading again Ancient Greek Philosophy for my own entertainment, I was astonished by the Epicurean view of pleasure, not quite what I was expecting, and its close correspondence to the notion of the Reinforcement Learning objective function. Following a Google search, I was even more astonished that nobody appeared to have made the link. Trapped in a cycle of urgent but non-important matters, I ended up writing up the talk as a text a couple of months later, during my summer vacation, which I further revised based on the kind feedback of Professors  L{\'u}cia Specia, Shimon Edelman and Neil Lawrence. I also thank Professor Lawrence for encouraging me to write this up in the first place, Professor Georg Struth for initial discussions on Epicurean Philosophy and Dr Tom Schaul for giving the thumps up. Special thanks to Professor Andy Barto for pointing out that Plato's ideas were also related to Reinforcement Learning, and for further discussions.

%(expert=\textgreek{>epa"'iwn}) 
%Protagoras relevant text 356c 357a,b

\bibliographystyle{unsrt}
\bibliography{biblio.bib}

\begin{thebibliography}{10}

\bibitem{Britannica}
Carlo Diano.
\newblock {\em Encyclopaedia Britannica}.
\newblock Encyclop{\ae}dia Britannica, Inc., 2010.

\bibitem{SEP}
Edward~N. Zalta, editor.
\newblock {\em The Stanford Encyclopedia of Philosophy}.
\newblock The Metaphysics Research Lab, Center for the Study of Language and
  Information, Stanford University, 2012.

\bibitem{Democritus}
Diogenes Laertius.
\newblock {\em Lives of Eminent Philosophers, Book IX, Chapter VII Democritus}.
\newblock Charpentier, Libraire-Editeur, {\url{http://remacle.org}}, 1847.

\bibitem{Girton}
Griton College.
\newblock {\url{https://www.girton.cam.ac.uk}}.

\bibitem{IEP}
Tim O'Keefe.
\newblock Epicurus.
\newblock The Internet Encyclopedia of Philosophy, ISSN 2161-0002
  {\url{http://www.iep.utm.edu}}, 2017.

\bibitem{Barto}
A.~G. Barto.
\newblock {\em Intrinsically motivated learning in natural and artificial
  systems}, chapter Intrinsic motivation and reinforcement learning.
\newblock Springer Berlin Heidelberg, 2013.

\bibitem{Epicurus}
Epicurus.
\newblock Letter to menoeceus.
\newblock Reproduced in Lives of Eminent Philosophers, Book X
  {\url{http://www.perseus.tufts.edu/}}.

\bibitem{EpicurusEN}
Epicurus.
\newblock Letter to menoeceus.
\newblock Translated in English by The Philosophical Garden
  {\url{http://www.philosophicalgarden.com}}.

\bibitem{SuttonBarto}
Richard~S. Sutton and Andrew~G. Barto.
\newblock {\em Introduction to reinforcement learning}.
\newblock Cambridge: MIT Press, 1998.

\bibitem{Plato}
Plato.
\newblock {\em Protagoras}.
\newblock The Collected Dialogues of Plato. Pantheon Books, a Division of
  Random House Inc. NY, 1963.

\bibitem{12GreeksRomans}
Carl~J. Richard.
\newblock {\em Twelve Greeks and Romans who Changed the World}.
\newblock Rowman \& Littlefield Publishers, 2003.

\bibitem{OnAristotle}
W.K.C. Guthrie.
\newblock {\em A History of Greek Philosophy: Volume 6, Aristotle: An
  Encounter}.
\newblock Cambridge University Press, 1981.

\bibitem{Aristotle}
Aristotle.
\newblock Nicomachean ethics.
\newblock Translated in English by W.D.Ross, {\url{http://classics.mit.edu}}.

\bibitem{ArtistotleMataphysics}
Vasilis Politis.
\newblock {\em Aristotle and the Mataphysics}.
\newblock Routledge, 2004.

\end{thebibliography}

\end{document}